# Multi-view learning for automatic classification of multi-wavelength auroral images

Qiuju Yang[1], Hang Su[1], Lili Liu[1], Yixuan Wang[1], and Ze-Jun Hu[2]

[1]School of Physics and Information Technology, Shaanxi Normal University, Xi'an, China,
[2] MNR Key Laboratory for Polar Science, Polar Research Institute of China, Shanghai, China.

*Abstract*—Auroral classification plays a crucial role in polar research. However, current auroral classification studies are predominantly based on images taken at a single wavelength, typically 557.7 nm. Images obtained at other wavelengths have been comparatively overlooked, and the integration of information from multiple wavelengths remains an underexplored area. This limitation results in low classification rates for complex auroral patterns. Furthermore, these studies, whether employing traditional machine learning or deep learning approaches, have not achieved a satisfactory trade-off between accuracy and speed. To address these challenges, this paper proposes a lightweight auroral multi-wavelength fusion classification network, MLCNet, based on a multi-view approach. Firstly, we develop a lightweight feature extraction backbone, called LCTNet, to improve the classification rate and cope with the increasing amount of auroral observation data. Secondly, considering the existence of multi-scale spatial structures in auroras, we design a novel multi-scale reconstructed feature module named MSRM. Finally, to highlight the discriminative information between auroral classes, we propose a lightweight attention feature enhancement module called LAFE. The proposed method is validated using observational data from the Arctic Yellow River Station during 2003-2004. Experimental results demonstrate that the fusion of multi-wavelength information effectively improves the auroral classification performance. In particular, our approach achieves state-of-the-art classification accuracy compared to previous auroral classification studies, and superior results in terms of accuracy and computational efficiency compared to existing multi-view methods.

## I. Introduction

AURORAS are colorful luminous phenomena caused by the collision of charged particles of the solar wind generated by solar activity, which accelerate towards the Earth along the magnetic lines of force and then collide with molecules or atoms in the Earth's atmosphere at the height of the ionosphere in the polar region [1]. The study of auroras is crucial for space physics as they serve as a powerful observational tool. The wavelength/color of an aurora correspond to the energy of the precipitating particles, whereas its brightness/intensity are determined by the number of precipitating particles. In addition, the morphological structure of an aurora reflects the spatial distribution of these particles [2]. Therefore, auroral morphology observed from the ground is essential for understanding magnetospheric dynamics, which has led to the research on ground-based auroral image classification [3], [4].

Ground-based observations of auroras show a wide variety of shapes, but a common classification scheme is lacking. In general, auroras can be classified into two main types: discrete auroras, which have structured shapes, and diffuse auroras, which are typically faint, blurred, and relatively homogeneous in appearance. Discrete auroras can be further subdivided into arcs and coronas based on their spatial morphology. Hu et al. [5] proposed a classification scheme consisting of arc auroras, drapery corona auroras, radial corona auroras and hot spot auroras based on the three-wavelength observations from the Yellow River Station (YRS) in Ny-Ålesund, Svalbard. This classification scheme has been widely accepted and used in subsequent studies [6], [7], [8].

All-sky CCD imagers (ASIs) are the most common ground-based auroral imaging devices [5], [9], [10]. For example, the YRS optical system consists of three ASIs equipped with narrow-band interferential filters centered at 427.8, 557.7, and 630.0 nm [5]. The auroral morphology at different wavelengths is not identical. Fig. 1(a) illustrates arc auroras, characterized by one or multiple discrete arcs that extend longitudinally. These arcs are long in the east-west direction and narrow in the north-south direction. In particular, the auroras at 557.7 nm exhibit more prominent and brighter arc structures, indicating a stronger emission at 557.7nm. Fig. 1(b) displays the drapery corona aurora, which is a faint ray-like corona aurora with an east-west orientation and weak emission at 557.7nm. The drapery corona aurora coexists with the diffuse aurora (the corona aurora appears at the poleward boundary of the diffuse aurora) and lacks a distinct outline. Fig. 1(c) reveals the radial corona aurora, which is characterized by radial rays emanating from the southeast, northwest, and north directions. These rays are particularly prominent, exhibiting strong excitation at 630.0 nm. Fig. 1(d)

This work was supported by the National Natural Science Foundation of China (grant 41504122), the Natural Science Basic Research Plan in Shaanxi Province of China (grant 2023-JC-YB-228), and the Open Fund of State Key Laboratory of Loess and Quaternary Geology (grant SKLLQGZR2201).

Q. Yang (corresponding author), H. Su, L. Liu, and Y. Wang are with the School of Physics and Information Technology, Shaanxi Normal University, Xi'an 710119, China (e-mail: yangqiuju@snnu.edu.cn; s1093553299@126.com;lilyliu@snnu.edu.cn; wyx1322588405@163.com).

Z.-J. Hu is with Key Laboratory for Polar Science, MNR, Polar Research Institute of China, Shanghai 200136, China (e-mail: huzejun@pric.org.cn).

exhibits the hotspot aurora, which has a complex and predominantly radial structures. It includes transiently brightening ray bundles, spots, and irregular patches, with rapid changes in brightness. The hotspot aurora exhibits strong emission at 427.8 nm, 557.7 nm, and 630.0 nm, featuring a distinct and bright spot structure, especially at 557.7 nm. Auroras show variations in brightness and morphology at different wavelengths. In particular, auroras recorded at 557.7 nm show the clearest and most well-defined morphological structures. In contrast, those observed at 630.0 nm show greater brightness, while the 427.8 nm auroras appear particularly faint. These variations are consistent across different observing hardware and station locations.

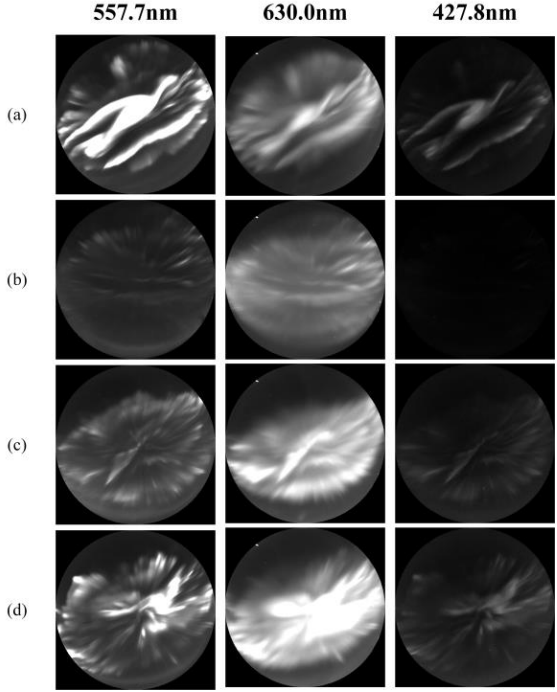

Fig. 1. Auroral observations at three wavelengths (a) arc, (b)drapery corona, (c) radial corona, (d) hotspots.

The classification of auroras encounters several challenges due to their inherent characteristics. First, there is considerable intraclass variation. The continuous evolution of auroras over time leads to substantial morphological variations within the same type. For instance, arc auroras can have different numbers, intensities, sizes, and shapes of arcs in their morphology, leading to significant differences in their appearance. Second, the differences between classes can be relatively small. Transitions between auroral types are often gradual, resulting in subtle differences between them. For instance, drapery coronas and radial coronas are often similar, as are radial coronas and hotspots. Third, the volume of auroral data is increasing significantly. For example, the three ASIs at YRS take millions of images per year, resulting in a substantial accumulation of auroral images. Fourth, the fusion of multi-wavelength data is essential for automatic auroral classification. This need arises from the unique variations in brightness and morphology that different auroral types exhibit at different wavelengths. However, there is very little research into automatic analysis of multi-wavelength auroras based on machine learning methods is lacking. Only when manually determining auroral types or studying auroral evolution have scientists combined auroral observations at different wavelengths [5], [11], [12].

In recent years, there has been a surge of research into automatic classification techniques for auroral images. Traditional machine learning methods for aurora classification have predominantly relied on manually designed features [13], [14], [15], [16], [17]. However, with the advent of artificial intelligence (AI) and the remarkable success of convolutional neural network (CNN)-based approaches in computer vision tasks, numerous CNN-based methods specifically tailored to auroral image classification have been developed and have shown promising classification performance [18], [19], [20], [21], [22], [23]. These methods typically leverage existing models for classification, resulting in improved classification accuracy but compromised classification efficiency. Considering the huge amount of auroral images captured annually, the focus of our research is to develop a fast and efficient auroral classification model. Furthermore, previous studies on automatic auroral analysis have mainly focused on the analysis of auroral images at a single wavelength, typically 557.7 nm. Consequently, the effective fusion of multi-wavelength auroral images is also a focus of this study.

To effectively exploit the information from multi-wavelength auroral images and achieve fast classification, this paper proposes a multi-wavelength lightweight fusion classification network, called MLCNet, based on the multi-view approaches. Multi-view here refers to different wavelengths in the same view. MLCNet aims to fully extract multi-scale features from different auroral wavelengths while incorporating the attention mechanism to suppress irrelevant information, thereby enabling effective auroral classification. First, we introduce the lightweight network LCTNet to extract features from images acquired at the three wavelengths. Second, we propose the multi-scale reconfiguration module, MSRM, to extract richer features from different receptive fields, thus improving the representation of multi-scale features. We then introduce the attention module, LAFE, which combines global and local contextual features to characterize the global contour and local texture information of the aurora. This module highlights important information that is crucial for classification. Finally, the features extracted from the three wavelength auroral images are fused to perform the auroral classification. To validate our proposed method, we conduct experiments on dayside auroral observations captured at YRS from 2003 to 2004. The experimental results demonstrate that our method achieves more accurate and efficient auroral classification. The main contributions of this work are as follows:
(1) This study presents a novel automatic auroral classification method (called MLCNet) that integrates multi-wavelength auroral information for the first time. It effectively combines auroral features extracted from different wavelengths, surpassing the performance of single wavelength classification. Furthermore, the analysis shows that auroral images at 630.0 nm are classified with higher accuracy than the more commonly used 557.7 nm images.
(2) To efficiently extract multi-wavelength auroral features and to enable fast classification of auroral images, a novel lightweight classification network architecture called LCTNet is proposed. This architecture is based on the ConvNeXt-Tiny network [24], which maintains the classification accuracy while significantly improving the efficiency.
(3) To address the multi-scale problem of auroras, the multi-scale reconfiguration module MSRM is introduced. This module efficiently extracts multi-scale features through hierarchical convolution, bridging the gap between different scales, reducing model parameters and computational cost, and achieving higher accuracy.
(4) To address the challenge of small differences between auroral classes, the paper proposes the LAFE attention module. It adaptively extracts the contextual information, resulting in more detailed auroral features and further improving the model's classification performance.
(5) Extensive experiments are conducted using auroral observations from Yellow River Station. The results show that the proposed MLCNet not only achieves state-of-the-art aurora classification accuracy, but also outperforms current multi-view methods in terms of both accuracy and computational efficiency.

## II. RELATED WORK

### A. Auroral Classification Based on Traditional Machine Learning

In the early stages of auroral image classification research, manual annotation of auroral data was the primary approach, requiring researchers to visually assess and classify the images [1], [25], [26]. However, due to the large volume of images captured by ground-based ASIs, this manual classification process became time-consuming and relied heavily on subjective judgment. Syrjäsuo et al. [14] were the first to apply computer vision methods to auroral image classification by utilizing shape skeleton analysis to classify images into four categories: arcs, patchy auroras, omega bands, and north-south structures. Since then, various machine learning-based approaches have been developed for auroral studies, including classification, segmentation, and retrieval.

Many studies have focused on auroral classification based on observations from YRS. Wang et al. [13] used local binary pattern operators and a block partition scheme to extract shape and texture information for auroral characterization. Yang et al. [27] proposed an auroral sequence characterization method based on hidden Markov models, incorporating inter-frame temporal information for classification. Zhong et al. [28] fused auroral MeanStd, Scale Invariant Feature Transform (SIFT), and Shape-Invariant Texture Index (SITI) using latent Dirichlet allocation to obtain topic features, and combined them with an SVM classifier for classification. Zhang et al. [29] employed a spectral clustering algorithm to divide multiple auroral feature representations into different groups, and used a label fusion method to fuse classification results from these representations.

However, these traditional machine learning-based auroral image classification techniques require manual design of feature extractors that can effectively capture relevant features such as shape, texture, and structure. The performance of these approaches depends heavily on the quality of the extracted features, which requires considerable experience and exploration time. Additionally, selecting appropriate classifiers is necessary to complete the classification process. Consequently, these methods do not provide a fully automated solution for auroral image classification.

### B. Auroral Classification Based on Deep Learning

In recent years, deep learning techniques have gained popularity in image processing due to their ability to autonomously learn and extract features. Clausen et al. [19] employed a pre-trained Inception-v4 neural network to automatically classify auroral images into six categories: clear/no aurora, cloudy, moon, arc, diffuse, and discrete. Yang et al. [21] utilized a combination of AlexNet and Spatial Transformer Networks to classify dayside auroras at the YRS into arc aurora, drapery aurora, radial aurora, and hotspot aurora. Kvammen et al. [30] compared six deep neural network architectures and found that the ResNet-50 architecture achieved the highest performance for classifying auroral images into seven major categories.

Transfer learning has also been applied to auroral image classification. Sado et al. [31] developed a transfer learning algorithm and tested 80 neural networks on the Oslo Aurora THEMIS (OATH) dataset. They used a deep learning model as a feature extractor and an SVM classifier to classify the extracted features. Guo et al. [32] compared different CNN architectures and number of layers for mesoscale aurora classification and found that ResNet-50 achieved the highest F1 score. Yang et al. [33] proposed a few-shot learning algorithm for auroral image classification, which allows fast recognition of new categories with limited labeled samples and reduces overfitting by using a cosine classifier.

Although these deep learning methods have shown good results in auroral image classification, they mainly focus on feature extraction from single wavelength auroral images using existing CNN models. They do not fully exploit the complementary information available in auroral images acquired at different wavelengths, nor do they effectively adapt the deep neural networks to the unique characteristics of auroral images. Furthermore, there has been limited research into the classification efficiency of these models, particularly in terms of computational speed.

### C. Deep Learning Based Multi-view Approaches

In recent years, the proliferation of multi-view data has become a significant trend on the internet, particularly in areas such as medical imaging and street view classification. The incorporation of multi-view analysis methods in these fields has also gained momentum. In the realm of medical image processing research, multi-view medical image analysis techniques are frequently employed to enhance the performance of computer-aided diagnosis (CAD) systems.

In mammography image analysis, each breast object is typically observed from two different views: the mediolateral oblique (MLO) view and the craniocaudal (CC) view. Numerous studies have explored mammography images using different views. Li et al. [34] utilized multi-view breast images as input and employed dilated convolutions to capture multi-scale breast information within a large "field of view". They extracted discriminative features from multiple breast views by taking the dilated convolution into a CNN architecture, and then connected the features from multiple views to several fully connected (FC) layers to produce the final output. Song et al. [35] simultaneously utilized multi-view and multi-modal images of the same breast as inputs. They extracted features from each input using existing network architecture and connected all feature branches, subsequently feeding them into FC layers to obtain the final output.

Among other multi-view medical image analysis methods, Luo et al. [36] extracted features of retinal lesions from multi-view fundus images. They employed an attention mechanism to explore the relationship between different views, calculated view weights, and fused them using a fusion function for diabetic retinopathy detection. Guo et al. [37] utilized multi-view colposcopic images as input and employed dual attention mechanisms, such as channel attention module CBAM (Convolutional Block Attention Module) and spatial attention module CA(Coordinate attention), to achieve feature selection and diagnosis of clinical colposcopic images.

In street-view classification analysis tasks, the combination of street-view and satellite imagery as multi-view representations has proven to be beneficial for joint classification. Hoffmann et al. [38] investigated model fusion techniques for satellite and street-view images to classify different urban building types, demonstrating the value of multi-modal data fusion. Barbierato et al. [39] combined high-resolution multispectral satellite imagery with data retrieved from the Google Street View database. They computed remote sensing metrics by incorporating Laser Imaging Detection and Ranging (LiDAR) data to generate indices at different heights relative to the ground. Ecological metrics were then computed using a pre-trained deep neural network based on proximate sensing. Chen et al. [40] utilized satellite imagery in conjunction with four angles of street-view images to learn spatial and multi-angle features of urban areas, aggregating these features and obtaining final prediction probabilities through a gated fusion module. These studies highlight the potential of the multi-view approach in street-view classification tasks.

In summary, traditional auroral image classification methods rely heavily on manually designed features, while CNN-based classification methods typically utilize only a single wavelength of auroral images, neglecting the complementary information present in different wavelengths. This highlights the need to develop lightweight classification models that can effectively leverage multi-wavelength auroral images for fast and accurate classification of auroras.

## III. METHOD

This paper proposes a lightweight network, MLCNet, for fast and effective classification of multi-wavelength auroral observations. MLCNet leverages multi-view learning techniques and domain knowledge of auroras. Fig. 2 illustrates the three main components of MLCNet: LCTNet, MSRM and LAFE attention module. LCTNet takes a minimalist approach to efficiently extract features from input images. MSRM improves the feature extraction capability of MLCNet by separating and splicing the multi-level feature maps for better extraction of auroral texture and structural features. The LAFE module effectively fuses semantically and scale-inconsistent features, focusing the network on the auroral regions of interest and eliminating interference from non-target regions. Finally, the acquired three-wavelength features are fused and the fused features are classified using Softmax.

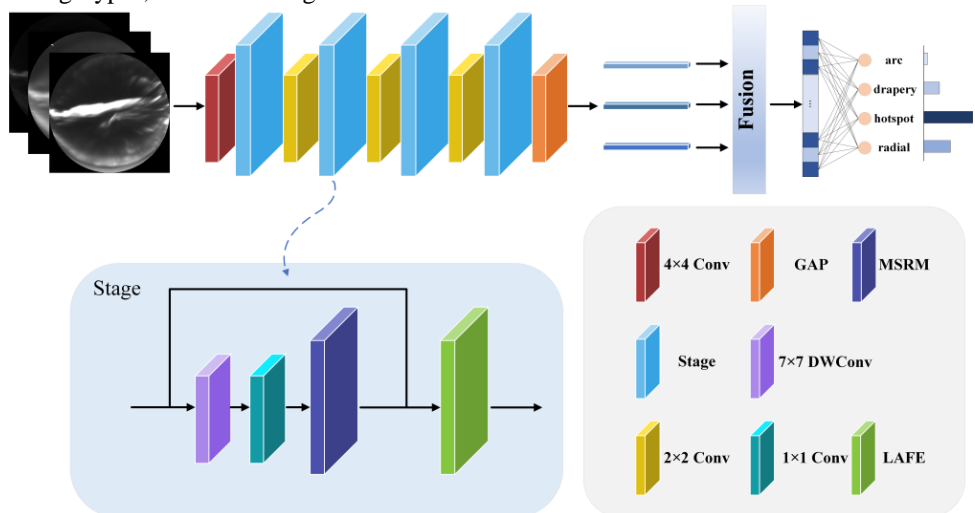

Fig. 2. Architecture of the proposed MLCNet. MLCNe contains three main components: LCTNet, MSRM and LAFE.

## A. Lightweight Feature Extraction Network LCTNet

The feature extraction network utilized in this study is based on the ConvNeXt-Tiny architecture [24], which is a state-of-the-art deep learning approach known for its high performance and compact model size. ConvNeXt-Tiny consists of four convolutional downsampling structures and a four-stage inverted bottleneck structure. Each stage of ConvNeXt-Tiny contains a different number of blocks: [3, 3, 9, 3].

Although ConvNeXt-Tiny has shown promising results in previous studies, it did not meet our need for efficient networking due to its large number of floating point operations (FLOPs), parameters (Params), and its long inference time (Infer). Consequently, the following modifications are made to improve the efficiency of ConvNeXt-Tiny:

(1) We adopt the concept of the inverted bottleneck structure but reduce the number of blocks in each stage to [1, 1, 1, 1]. This modification significantly reduces the number of Params by approximately two-thirds and the FLOPs by nearly three-quarters.
(2) We keep the original 7×7 depth-wise convolution (DWConv) in the block, but replace the first 1×1 convolution layer with a 1×1 Group convolution with fewer parameters.
(3) To improve the stability of the training, we introduce a separate convolutional downsampling and batch normalization (BN) layer between the stages where the spatial resolution changes, and incorporate the BN along with the rectified linear unit (ReLU) activation function into the network training process.

Fig. 3 details the block changes. By implementing these improvements, our modified lightweight ConvNeXt-Tiny network (referred to as "LCTNet") increases efficiency without compromising performance.

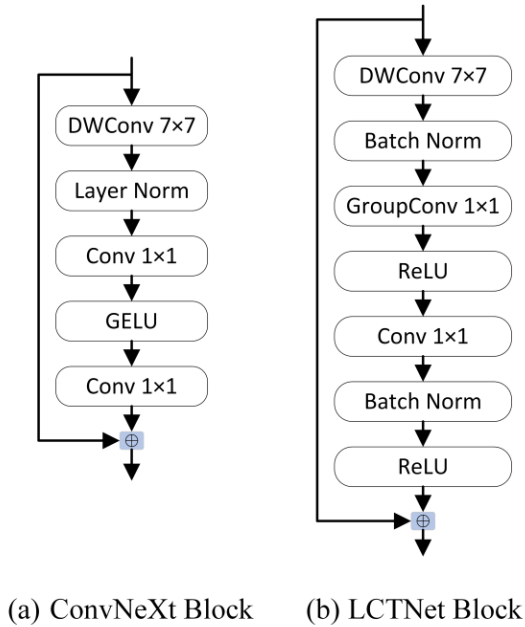

Fig. 3. Block designs for a ConvNeXt (a) and an LCTNet (b).

## B. Multi-scale Reconfiguration Module MSRM

Fig. 4(a) illustrates the schematic diagram of MSRM. In order to efficiently acquire auroral multiscale features, we take inspiration from HS-ResNet [41] and use Hierarchical-Split Block (HSBlock) to extract multiscale features at a fine-grained level. The HSBlock uses a 3×3 convolution operation at each level, which only processes a subset of the feature map channels, limiting the number of features acquired. In response, we developed the Re-parameterization Convolution block (RECblock), shown in Figure 4(b), as an alternative to the 3×3 convolution. RECblock acquires features by stacking deep convolutions (DWConv) with reduced parameters and lower computational cost. This approach enlarges the receptive field of each network layer and exploits the variability of the receptive field at the fine-grained level to capture both detailed and global features. Furthermore, to maintain the lightweight nature of the model, RECblock can be adapted using the parameter reconstruction concept of RepVGG [42]. Recognizing that the final stage of feature channel splicing in HSBlock does not facilitate effective information exchange between different groups of feature maps, we introduced the channel shuffle operation. This operation allows the flow of feature information between different groups.

MSRM starts by dividing the input features $X \in \mathbb{R}^{B \times C \times H \times W}$ into five groups, each with an equal number of channels, represented as $x_i$. Among these groups, only the first group of feature maps, $x_1$, is directly fed into the final output. The second group of feature maps, $x_2$, is passed through the RECblock, generating two sets of sub-feature maps, $t_{2,1}$ and $t_{2,2}$. One group of sub-feature maps, $t_{2,1}$, is spliced into the final output, while the other group, $t_{2,2}$, is spliced into the third group of feature maps. This iterative process continues until the splicing results of all five groups of feature maps are obtained. The resulting output features can be expressed as follows:

$$t_i = \begin{cases} x_i, & i = 1 \\ E_i(x_i \oplus t_{i-1,2}), & 1 < i <= 5 \end{cases} \quad (1)$$

where $E_i$ represents the RECblock and $t_i$ denotes the output feature mapping of $E_i$. The channels are then divided into sub-channels by the channel shuffle operation, where they are grouped into different subgroups. Finally, a $1 \times 1$ convolution is employed to merge the resulting blended feature maps.

This approach enables the final output features to encompass different scales of receptive fields from various groups of feature maps. Feature maps closer to the front involve fewer convolutions, have smaller receptive fields, and focus on capturing detailed information. Conversely, feature maps located towards the back have more convolutions and larger receptive fields, and emphasize global information [41]. Fig. 4(b) shows the RECblock, which consists of a training phase and an inference phase. During the training phase, a two-branch structure is employed. The feature $x_i$ is processed through a $3 \times 3$ DWConv + BN branch and a $1 \times 1$ DWConv + BN branch, resulting in output features denoted as $t_i$, which are then combined through summation and fusion. Although this structure offersperformance advantages, it significantly increases memory consumption and reduces the speed of model inference. To address these challenges, a parameter reconstruction technique is introduced at the inference stage. This technique reconfigures the $3 \times 3$ DWConv + BN and $1 \times 1$ DWConv + BN layers from the training stage into a unified $3 \times 3$ DWConv layer. This process involves converting the $1 \times 1$ DWConv into a $3 \times 3$ DWConv with zero-filling and merging the DWConv of the dual branch together with BN into a $3 \times 3$ DWConv layer with bias. Finally, the convolution kernel and bias obtained by the double branch were added separately. By implementing this approach, the originally trained dual branch structure is effectively transformed into a single branch configuration with a single $3 \times 3$ DWConv operation.

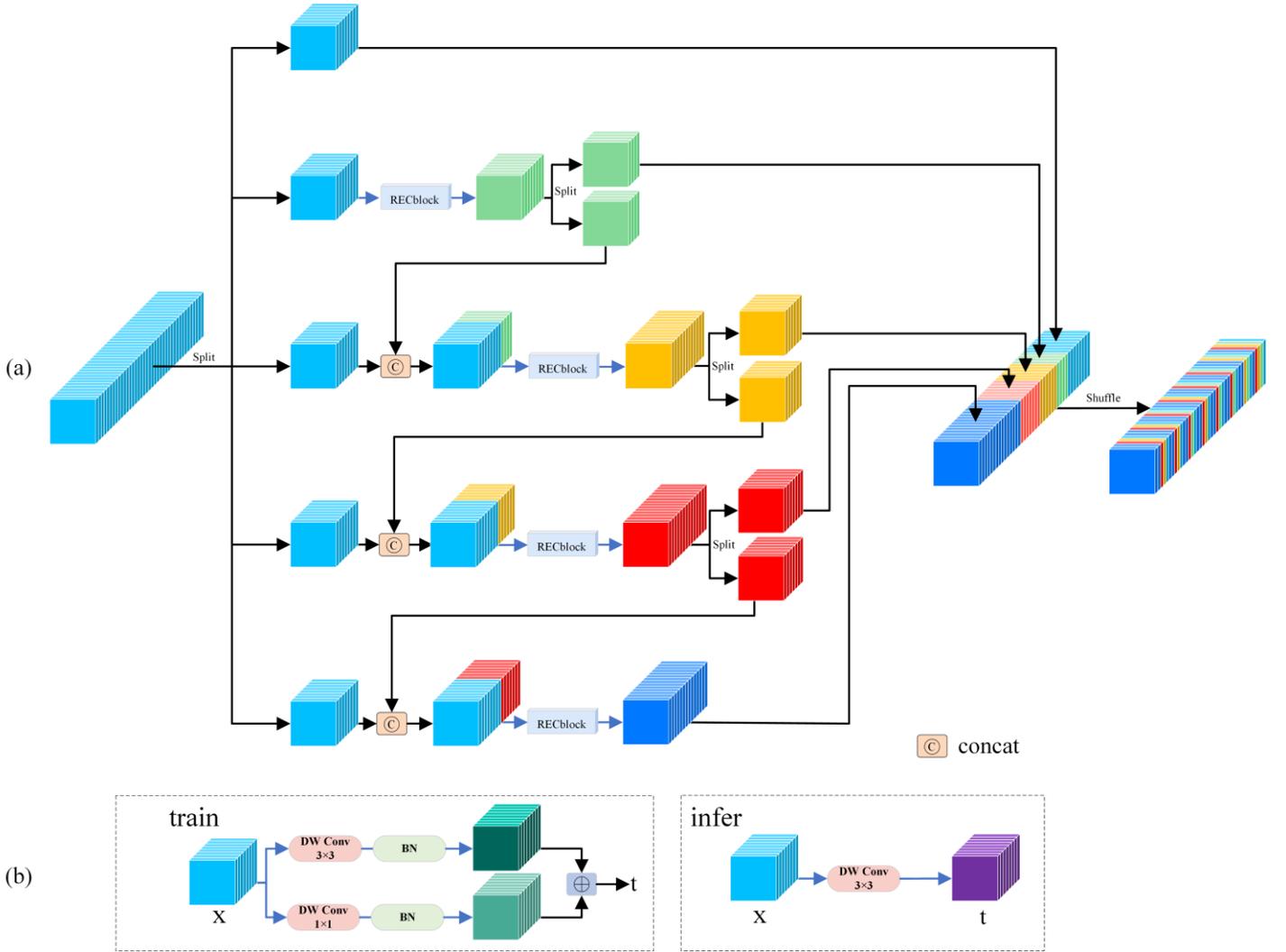

Fig. 4. (a) Structure of the Multi-scale Reconfiguration Module (MSRM). (b) The training and inference phases of RECblock. The 3×3 DWConv+BN and 1×1 DWConv+BN layer from the training phase are reconstructed into a 3×3 DWConv layer in the inference phase.

### C. Lightweight Attention Feature Enhancement Module LAFE

To aggregate multi-scale contextual features and integrate local and global information extracted by the network to better capture differences in auroral categories, we developed the LAFE module. As illustrated in Fig. 5, LAFE consists of two parts: local feature channel attention and global feature channel attention. In contrast to the AFF module [43], we use DWConv to incorporate scale changes into the features by replacing the initial two $1 \times 1$ point convolutions with a single $1 \times 1$ DW convolution. This strategy ensures model efficiency and prevents the loss of feature maps from different groups obtained by the MSRM group convolution due to the changing the change of the channel. To improve the model's capacity for nonlinear mapping and to better account for the complex correlations between channels, we introduced ReLU in the final summation stage.

Given an input feature $X \in \mathbb{R}^{B \times C \times H \times W}$ with skip-connections and an identity input feature $Y \in \mathbb{R}^{B \times C \times H \times W}$, they are first summed to obtain the feature map $I \in \mathbb{R}^{B \times C \times H \times W}$. The feature map is generated by the following calculation:

$$I = X \oplus Y \quad (2)$$

here, $\oplus$ denotes the broadcasting addition.

Next, the global feature channel attention, denoted as $G(I) \in \mathbb{R}^{B \times C \times 1 \times 1}$, and the local feature channel attention, denoted as $L(I)$, are computed separately. The calculation for the global feature channel attention is as follows:

$$G(I) = \beta\left(\delta\left(\beta\left(Conv(g(I))\right)\right)\right) \quad (3)$$

where $g(I) = \frac{1}{H \times W} \sum_{i=1}^{H} \sum_{j=1}^{W} I_{[:,:,i,j]}$ represents the Global Average Pooling (GAP), Conv denotes the $1 \times 1$ DWConv, $\delta$ is the ReLU activation function, and $\beta$ is the BN. The local feature channel attention, $L(I)$, differs from $G(I)$ by excluding one step of GAP operation. It is worth noting that $L(I)$ retains the same shape as the input feature, preserving and emphasizing fine details within the lower-level features.

With the global feature channel attention $G(I)$ and the local feature channel attention $L(I)$, the LAFE attention module performs feature detection using the following formulas:

$$X' = X \otimes \sigma(G(I) \oplus L(I)) \quad (4)$$
$$Y' = Y \otimes \left(1 - \left(\sigma(G(I) \oplus L(I))\right)\right) \quad (5)$$
$$Out = X' + Y' \quad (6)$$

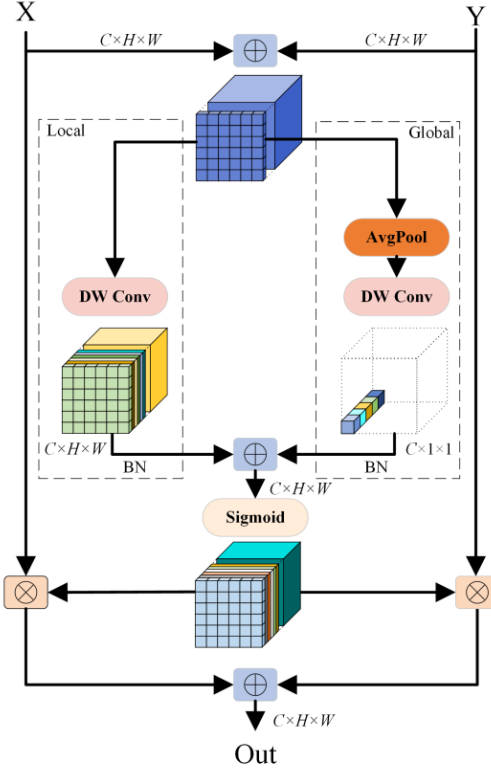

Fig. 5. Structure of the Lightweight Attention Feature Enhancement Module (LAFE). The local branch uses DW convolution directly to extract the attention from local features. The global branch uses Global Avg Pooling and DWConv to extract the attention from global features.

where $\sigma$ represents the Sigmoid activation function and $\otimes$ denotes element-by-element multiplication. The output values obtained by summing the features $I \in \mathbb{R}^{B \times C \times H \times W}$ are normalized between 0 and 1 using the Sigmoid activation function. The input features $X$ and $Y$ are given adaptive weights through (4) and (5), respectively. During the training process, the weighting parameters are determined to allow the network to perform a soft selection or weighted averaging between $X$ and $Y$ [43].

### D. Multi-wavelength Fusion Classification

In MLCNet, the multi-wavelength fusion classification leverages the combination of different wavelengths of auroral observations to gain a more comprehensive understanding of the aurora. First, a feature map $U_i$ is generated for each wavelength of the input image data. Each feature map $U_i$ is then subjected to adaptive average pooling [24], resulting in a corresponding feature vector $T_i$. This process is repeated for each wavelength, generating a set of feature vectors for each view. To create an integrated feature vector, a max operation is performed element-wise on the feature vectors from the different wavelengths. This produces a single feature vector that captures the most predominant features across the wavelengths. Finally, the FC layer utilizes the integrated feature vector to generate the final prediction Q. This can be represented using (7) as:

$$Q = FC(max(T_i)) = FC\left(max(Adapool(U_i))\right) \quad (7)$$

## IV. EXPERIMENTS AND RESULTS

### A. Dataset

The auroral images used in this study were captured at three distinct wavelengths (427.8 nm, 557.7 nm, and 630.0 nm) using three ASIs at the Arctic Yellow River Station (YRS) during the winter of 2003-2004. YRS is located at 76.24° magnetic latitude (MLAT) in Ny-Ålesund, Svalbard, providing favorable conditions for observing the aurora. The ASIs at the station allow continuous monitoring of the aurora with high spatial resolution, which is carried out from December to February each year. The YRS optical system consists of three ASIs equipped with narrow-band interferential filters centered at 427.8, 557.7 and 630.0 nm. The optical lens configuration includes a fish-eye lens with a FOV of 180°, a relay lens, a filter lens (interferential filters centered at 427.8, 557.7 and 630.0 nm with width of 2 nm, respectively) and a focus lens [5]. The full width at half maximum (FWHM) of the filter is 2 nm. The ASI captures images at a resolution of $512 \times 512$ with a temporal resolution of 10s, while the exposure and readout times are 7 and 3s respectively.

Based on the observations obtained from the three ASIs at YRS, Hu et al. [5] proposed a classification scheme for dayside auroras, which includes four main categories: arcs, drapery coronas, radial coronas, and hot spots. In this study, the auroral images were classified according to this scheme. The dataset used for classification consists of 8001 images captured from December 2003 to February 2004, including 3934 arc images, 1786 drapery images, 784 radial images, and 1497 hotspot images.

To ensure consistency and facilitate further processing, the images in the dataset underwent preprocessing steps similar to those described in the previous work [13]. These include systematic noise subtraction, intensity scaling, rotation to align the top of the image with the northern position, and cropping to remove extraneous light sources and areas of distortion. The final cropped auroral images were resized from $512 \times 512$ pixels to $440 \times 440$ pixels.

*B. Implementation Details*

*1) Data Augmentation:*

The input auroral images were first resized to $224 \times 224$ pixels before being fed into the neural network. The dataset was then divided into a training set and a test set in a 6:4 ratio. For each wavelength of auroral images, the training set consisted of 4803 training images and 3198 validation images. To counteract potential overfitting, various data augmentation techniques such as scaling, cropping and rotation were applied to the training images. This augmentation process helps the network to capture different variations and patterns within the data. Table I lists the training and test samples for each class.

TABLE I
AURORA CATEGORIES AND THE NUMBER OF SAMPLES PER CATEGORY.

| No. | Class | Train | Test |
|---|---|---|---|
| 1 | arc | 2361×3 | 1573×3 |
| 2 | drapery | 1072×3 | 714×3 |
| 3 | hotspot | 471×3 | 313×3 |
| 4 | radial | 899×3 | 598×3 |
| Total | | 4803×3 | 3198×3 |

*2) Experimental Details:*

The model was trained on a server equipped with an Intel i7-6700 CPU clocked at 3.40 GHz, 128 GB of memory, and a 12 GB NVIDIA GeForce RTX 1080Ti GPU. The server was running the Ubuntu 18.04.5 LTS operating system. The training process consisted of 150 epochs with a batch size of 16. We employed the PyTorch framework and utilized the AdamW optimizer for exponential learning rate decay. Cross-entropy loss was employed for both training and optimization. The initial learning rate was set to 0.0001 with a momentum of $5 \times 10^{-2}$. The learning rate was gradually reached within one epoch using linear warm-up and subsequently decayed exponentially.

*3) Evaluation Metrics:*

In order to evaluate the performance of the network, accuracy (Acc), average of accuracy values for all categories (Avg_acc) and F1 score were used to measure the recognition accuracy of the network. These metrics are defined as

$$\text{Accuracy} = \frac{TP + TN}{TP + TN + FP + FN} \quad (8)$$

$$F1 = \frac{2 \times \text{Recall} \times \text{Precision}}{\text{Recall} + \text{Precision}} \quad (9)$$

$$\text{Precision} = \frac{TP}{TP + FP} \quad (10)$$

$$\text{Recall} = \frac{TP}{TP + FN} \quad (11)$$

here TP means true positives, FP is false positives, TN represents true negatives, and FN is false negatives. Accuracy measures the overall effectiveness of the network, while Avg_acc assesses the classification performance on each category. The F1 score combines precision and recall to provide an overall evaluation criterion. In addition, FLOPs, Params and inference time were considered to measure the complexity and speed of the networks, allowing for a more comprehensive performance comparison between models.

*C. Ablation Experiment*

*1) Effect of Input Size*

To investigate the effect of input size on model performance, we conducted ablation experiments using the original image size of $440 \times 440$ and the cropped image of $224 \times 224$ as inputs, respectively, as shown in Table II. It is evident that both ConvNeXt-Tiny and LCTNet yield higher F1 values when the input size is set to $224 \times 224$. This improvement is due to the fact that the backbone network (ConvNext) was originally designed for an input size of $224 \times 224$. It is worth noting that increasing the input size significantly increases FLOPs and inference time. Therefore, we opt to use the cropped $224 \times 224$ images for the subsequent experiments.

*2) Effectiveness of different wavelengths*

We conducted a comparison of network performance using different wavelengths as model inputs in three models: the original ConvNeXt-Tiny, our backbone network LCTNet, and the final model of this paper, MLCNet, with the aim of validating the effectiveness and contribution of multiple wavelengths.

The results of the ablation experiments are presented in Table III, where the following observations are made for each model:

Using an auroral image with a wavelength of 630.0 nm as input gives better performance than using an auroral image with a wavelength of 557.7 nm, which is commonly used in previous studies. The performance of the model using 427.8 nm as input is the least favorable of the three wavelengths;

Using 630.0 nm and 557.7 nm auroral images as a two-wavelength input significantly improves model performance compared to single-wavelength inputs. Moreover, the addition of 427.8 nm as a three-wavelength input results in a slight improvement in model performance. This demonstrates the feasibility of using multi-wavelength images for classification tasks.

Compared to ConvNeXt-Tiny, our LCTNet achieves competitive classification accuracy (Acc) and average accuracy (Avg_Acc) when using single and multi-wavelength inputs, despite utilizing only 1/4 of the model parameters and 1/5 of the FLOPs, albeit with slightly lower results for the 630.0nm wavelength input.

Similarly, when compared to ConvNeXt-Tiny, the Acc and Avg_Acc accuracies for the proposed MLCNet are significantly improved using only 1/6 of the model parameters and 1/8 of the FLOPs. This improvement is observed in both the single-wavelength input and three-wavelength input scenarios. In particular, for three-wavelength data, the best classified Acc and Avg_Acc for MLCNet are 96.97% and 95.86% respectively, outperforming ConvNeXt-Tiny by 1.10% and 1.37% respectively. For single wavelength data, the classification accuracy improves significantly from 89.06% to 94.56% at 557.7 nm. At 427.8 nm, the improvement is even more significant, with classification accuracy increasing from 55.66% to 90.93%.

TABLE II
ABLATION EXPERIMENT OF THE INPUT SIZE

| Method | Size | Evaluation index | | | | | |
|---|---|---|---|---|---|---|---|
| | | Acc | Avg_Acc | F1 | Params (M) | FLOPs (G) | Infer (ms) |
| ConvNeXt-Tiny | 440×440 | 94.22% | 92.06% | 92.17% | 28.57 | 49.73 | 38.26 |
| | 224×224 | **95.87%** | **94.49%** | **94.47%** | | 13.37 | **12.19** |
| LCTNet | 440×440 | **95.65%** | **94.37%** | 94.06% | 6.60 | 9.36 | 7.33 |
| | 224×224 | **95.65%** | 93.97% | **94.20%** | | 2.51 | **2.55** |

TABLE III
ABLATION EXPERIMENT OF DIFFERENT WAVELENGTHS

| Method | Data | Acc | Avg_Acc | Params(M) | FLOPs(G) |
|---|---|---|---|---|---|
| ConvNeXt-Tiny | 557.7 | 89.06% | 85.46% | 28.57 | |
| | 630.0 | 94.28% | 92.69% | | 4.46G |
| | 427.8 | 55.66% | 34.96% | | |
| | Two-wavelength | 95.84% | **94.59%** | | 8.92G |
| | Three-wavelength | **95.87%** | 94.49% | | 13.37G |
| LCTNet | 557.7 | 89.81% | 86.25% | 6.60M | |
| | 630.0 | 91.92% | 89.31% | | 0.84G |
| | 427.8 | 69.67% | 57.14% | | |
| | Two-wavelength | 95.58% | 93.67% | | 1.67G |
| | Three-wavelength | **95.65%** | **93.97%** | | 2.51G |
| MLCNet (LCTNet + MSRM + LAFE) | 557.7 | 94.56% | 92.42% | 4.57M | |
| | 630.0 | 95.84% | 94.19% | | 0.55G |
| | 427.8 | 90.93% | 87.82% | | |
| | Two-wavelength | 96.68% | 95.57% | | 1.10G |
| | Three-wavelength | **96.97%** | **95.86%** | | 1.65G |

To further clarify the differences in classification between the models for 630.0 nm and 557.7 nm wavelengths, we generated confusion matrices for the three models, as shown in Fig. 6. It can be seen that when the input image is changed from 557.7 nm to 630.0 nm, all three models show improved performance for all four auroral types. Notably, the improvement is particularly pronounced for the two coronal auroras, drapery and radial, compared to the arc and hotspot auroras. This discrepancy arises because arc and hotspot auroras have well-defined structural features and perform well at all wavelengths. In contrast, drapery and radial auroras emphasize textural features, with weak emission at 557.7 nm and strong emission at 630.0 nm.

*3) Effectiveness and Generalizability of Modules*

To further verify the generalization and effectiveness of the designed MSRM and LAFE modules, we introduced these two modules as well as HSBlock and AFF to ConvNeXt-Tiny and LCTNet for ablation experiments. The results presented in Table IV show that the incorporation of MSRM and LAFE modules leads to significant improvements in classification accuracy and F1 value for both ConvNeXt-Tiny and LCTNet.

When the MSRM module was integrated into the ConvNeXt-Tiny network, the number of parameters and FLOPs were significantly reduced by 12.61M and 5.98G, respectively. This integration resulted in improvements of 0.14% in Acc_avg and 0.37% in F1. On the other hand, adding HSBblock to the ConvNeXt-Tiny network drastically increased the number of parameters and FLOPs by 34.54M and 16.81G respectively, while improving Acc_avg and F1 by only 0.07% and 0.14% respectively.

Adding the LAFE module to the network resulted in a marginal increase of 0.04M parameters and 0.03G FLOPs, while achieving superior improvements of 0.40% in Acc and 0.45% in Acc_avg. This indicates that the model improvements achieved by the LAFE module come at a lower computational cost compared to the addition of AFF.

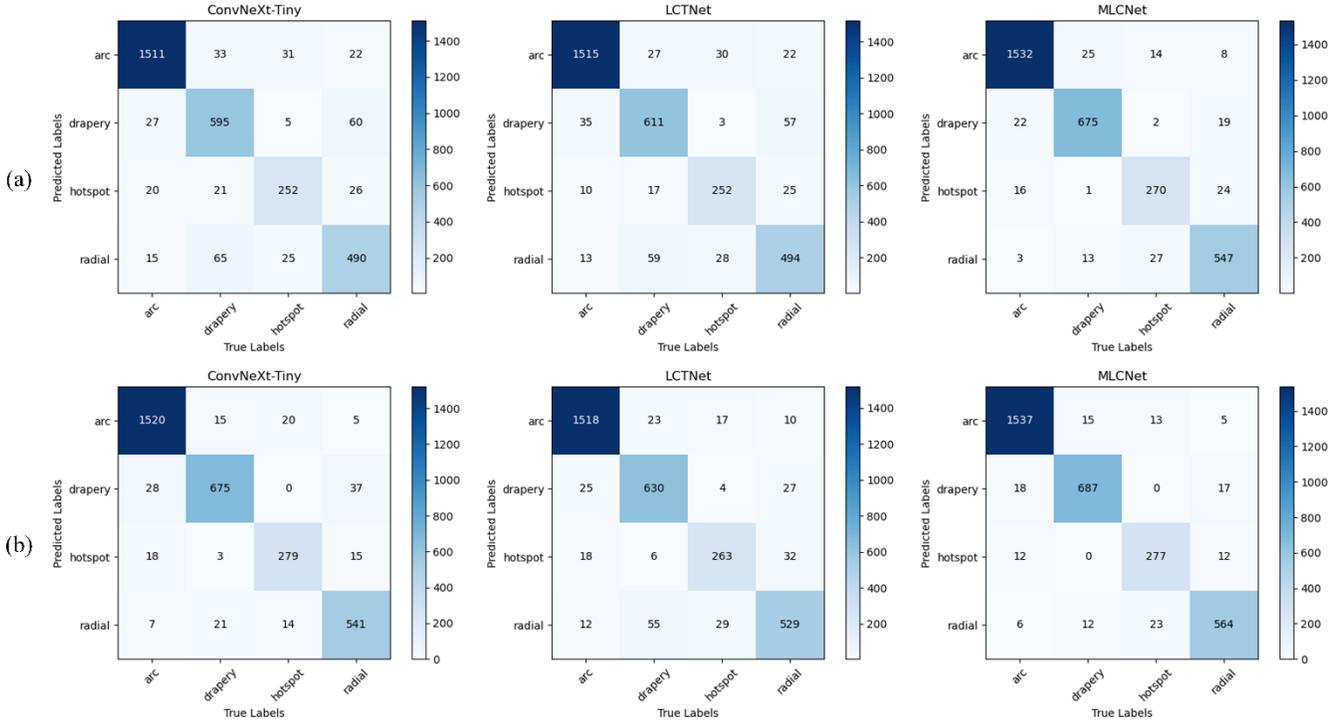

Fig. 6. Confusion matrix of the three neural networks on the test set. (a) 557.7nm images as input, (b) 630.0nm images as input

TABLE IV
ABLATION EXPERIMENT OF THE PROPOSED MODULES

| Method | Modules | | | | Evaluation index | | | | | |
|---|---|---|---|---|---|---|---|---|---|---|
| | HSBlock | MSRM | AFF | LAFE | Acc | Avg_Acc | F1 | Params (M) | FLOPs (G) | Infer (ms) |
| ConvNeXt-Tiny | | | | | 95.87% | 94.49% | 94.47% | 28.57 | 13.37 | 12.19 |
| | √ | | | | 95.96% | 94.56% | 94.70% | 63.11 | 30.18 | 22.10 |
| | | √ | | | 96.19% | 94.63% | 94.84% | **15.96** | **7.39** | **18.55** |
| | | | √ | | 96.18% | 94.71% | **95.02%** | 31.84 | 14.19 | 20.60 |
| | | | | √ | **96.27%** | **94.94%** | 94.97% | **28.61** | **13.40** | **18.73** |
| | | √ | | √ | **96.31%** | **95.26%** | **95.08%** | 16.00 | 7.41 | 24.59 |
| LCTNet | | | | | 95.65% | 93.97% | 94.20% | 6.60 | 2.51 | 2.55 |
| | √ | | | | 95.93% | 94.18% | 94.26% | 15.98 | 6.72 | 5.18 |
| | | √ | | | 96.21% | 94.61% | 94.88% | **4.56** | **1.64** | **4.23** |
| | | | √ | | 96.34% | 94.55% | 95.07% | 7.37 | 2.70 | 4.12 |
| | | | | √ | **96.46%** | **94.98%** | **95.21%** | 6.60 | 2.53 | **3.30** |
| | | √ | | √ | **96.97%** | **95.86%** | **95.95%** | 4.57 | 1.65 | 5.50 |

In summary, the proposed MSRM and LAFE modules outperform HSBlock and AFF in terms of accuracy, number of parameters and inference time. These positive results were consistently observed when these modules were applied to the LCTNet network, further confirming their effectiveness and generalizability.

Furthermore, when both MSRM and LAFE modules were incorporated into the ConvNeXt-Tiny network, the number of parameters and FLOPs were reduced by 12.57M and 5.96G respectively, while improvements of 0.77% in Acc_avg and 0.61% in F1 were achieved. Despite the doubling of model inference time with the inclusion of these two modules, it is important to note that our method still maintains the most competitive overall inference speed. This means that even with the increased computation time, our method maintains a competitive overall inference speed.

*4) Fusion Approaches*

The impact of different fusion approaches on our proposed method was investigated through four sets of experiments. The results are presented in Table V. It can be seen that the fusion approach employing the max operation on features at the final stage achieves the highest classification accuracy and computational efficiency.

TABLE V
COMPARISON OF DIFFERENT FUSION METHODS.

| Method | Acc | Avg_Acc | Params(M) | FLOPs(G) |
|---|---|---|---|---|
| add | 96.06% | 94.84% | 4.57M | 1.65G |
| min | 96.50% | 95.47% | 4.57M | 1.65G |
| con | 96.62% | 95.57% | 6.10M | 1.65G |
| max | **96.97%** | **95.86%** | **4.57M** | **1.65G** |

*D. Network Visualization*

To enhance the interpretability of our proposed modules, we employed Class Activation Mapping (CAM) to visualize the regions of interest for each model during the classification of auroral images at different wavelengths. Fig. 7 illustrates the visualization results.

For both arc auroras (Fig. 7(a)-(d)) and hotspot auroras (Fig. 7(e)-(h)), it can be seen that the model focuses on the auroral location using each wavelength of data. The regions of interest for the wavelengths of 557.7 nm and 427.8 nm exhibit a larger overlap, while there is a greater difference for the wavelength of 630.0 nm. This discrepancy is due to the fact that the wavelengths of 557.7 nm and 427.8 nm primarily differ in brightness, whereas 630.0 nm differs more significantly in morphology. By merging the results, it is possible to focus more precisely on the region where the auroras of interest are located. In the MSCNet (LCTNet + MSRM) visualization results (Fig. 7(e) and (f)), some attention is drawn to the uninterested auroras in the lower left. However, after fusion, the network no longer focuses on this area.

For easily classified arc auroras (Fig. 7(a)-(d)), all three models focus on the arc auroras, and the fused visualization results further refine the localization. On the other hand, for hotspot auroras that are prone to misclassification (Fig. 7(e)-(h)), the regions that LCTNet focuses on are not as precise, for example, they focus on the lower left auroras. With the addition of the modules MSRM and LAFE, the network (MSCNet and MLCNet respectively) become increasingly focused on the target region with increasing precision. These modules help eliminate redundant information that does not require attention. In particular, the fused visualizations demonstrate that the auroral regions change from irregular blocks to more precise contours.

In conclusion, the CAM visualization provides insight into the regions of interest for each model and demonstrates the improvement in auroral location accuracy achieved through multi-wavelength fusion and module additions.

*E. Comparison with Existing Auroral Classification Methods*

Previous research in the classification of auroral images primarily employed basic CNN models and traditional classifiers, which have their inherent limitations. In this study, we propose the MLCNet method, which addresses these limitations by combining multi-wavelength data and improving the basic model used in earlier studies. The above experimental results demonstrate the superiority of the classification of multi-wavelength auroral images over single-wavelength images, with three wavelength auroral images proving to be the most effective input choice. Table VI shows only the classification results of the different methods when three-wavelength auroral images are used as input. It is noteworthy that MLCNet surpasses previous methods in terms of accuracy and F1 score, while at the same time overcoming the problems of an excessive number of parameters and slow inference speed. Specifically, in comparison to the second ranked ConvNeXt-Tiny model, MLCNet achieves higher accuracy in both the Acc and Avg_Acc metrics, outperforming it by 1.10% and 1.37% respectively. MLCNet also outperforms ConvNeXt-Tiny on the F1 metric by 1.48%. In addition, MLCNet exhibits a faster inference speed, more than twice that of ConvNeXt-Tiny. The outstanding classification performance of MLCNet, combined with relatively low number of FLOPs and parameters, can be attributable to the integration of a multi-scale module and an attention module. This design enables MLCNet to achieve high classification accuracy with reduced inference latency. In conclusion, our results clearly demonstrate that the proposed MLCNet method represents a significant advance over previous approaches to auroral image classification, offering higher accuracy and faster inference times.

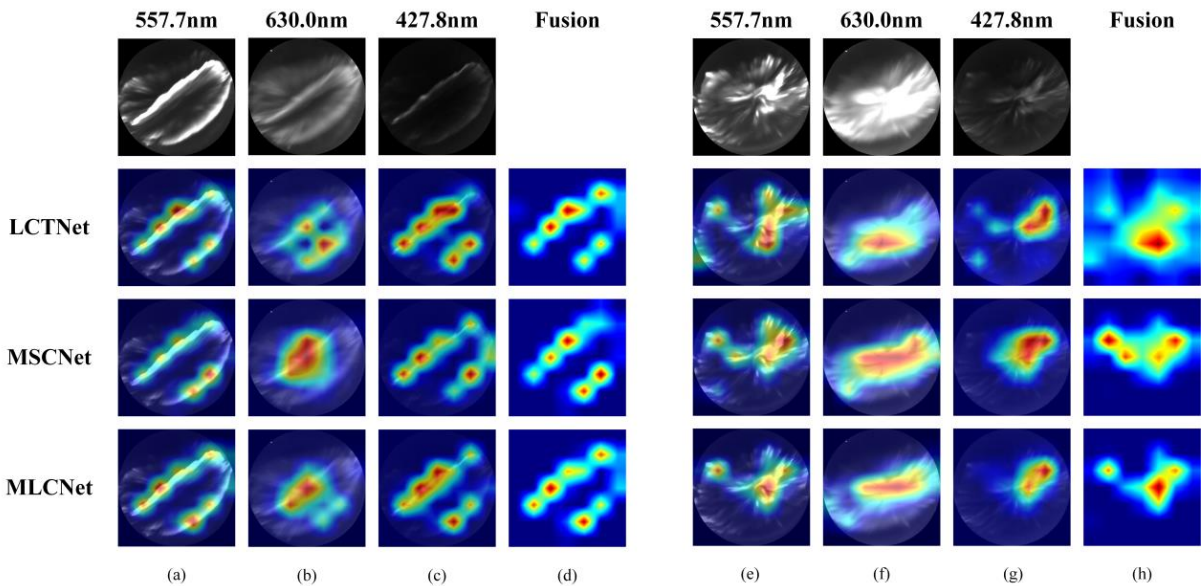

Fig. 7. Visualization of attention maps of the last convolutional layer of the last stage from different networks. Red color indicates higher attention values and blue color indicates lower values.

Fig. 8 illustrates the performance of MLCNet (highlighted by the red circle) compared to other models, taking into account performance metrics such as parameters, FLOPs and inference time. A recent study [50] highlighted that the inference speed for a batch size of 1 is closely related to the network depth and is independent of the number of parameters. This observation suggests that the use of simple and shallow networks is promising for real-time processing. It is clear from Fig. 8 that MLCNet significantly improves model effectiveness with minimal increase in FLOPs, inference latency and the number of parameters. This results in a commendable balance between speed and accuracy, significantly outperforming other comparative methods. These results highlight the great potential of MLCNet in the field of auroral image classification.

TABLE VI
CLASSIFICATION RESULTS OF DIFFERENT CNN MODELS

| Method | Acc | Avg_Acc | F1 | Params(M) | FLOPs(G) | Infer(ms) |
|---|---|---|---|---|---|---|
| AlexNet [30][44] | 89.06% | 85.81% | 85.97% | 16.63 | 0.90 | **0.88** |
| ResNet18 [30][45] | 95.62% | 94.28% | 94.22% | 11.69 | 5.53 | 3.52 |
| ResNet50 [32][45] | 93.96% | 91.93% | 92.02% | 25.56 | 12.33 | 7.92 |
| DenseNet121 [31][46] | 95.22% | 93.25% | 93.55% | 7.98 | 8.59 | 17.62 |
| ShuffleNetV2_2.0 [47] | 94.43% | 92.31% | 92.51% | 7.40 | 1.79 | 7.15 |
| MobileNetV3-Small [48] | 95.03% | 93.65% | 93.15% | **2.94** | **0.19** | 7.43 |
| EfficientNetv2_s [49] | 93.90% | 91.11% | 91.50% | 21.46 | 25.34 | 20.62 |
| ConvNeXt-Tiny [23][24] | 95.87% | 94.49% | 94.47% | 28.57 | 13.37 | 12.19 |
| MLCNet | **96.97%** | **95.86%** | **95.95%** | 4.57 | 1.65 | 5.50 |

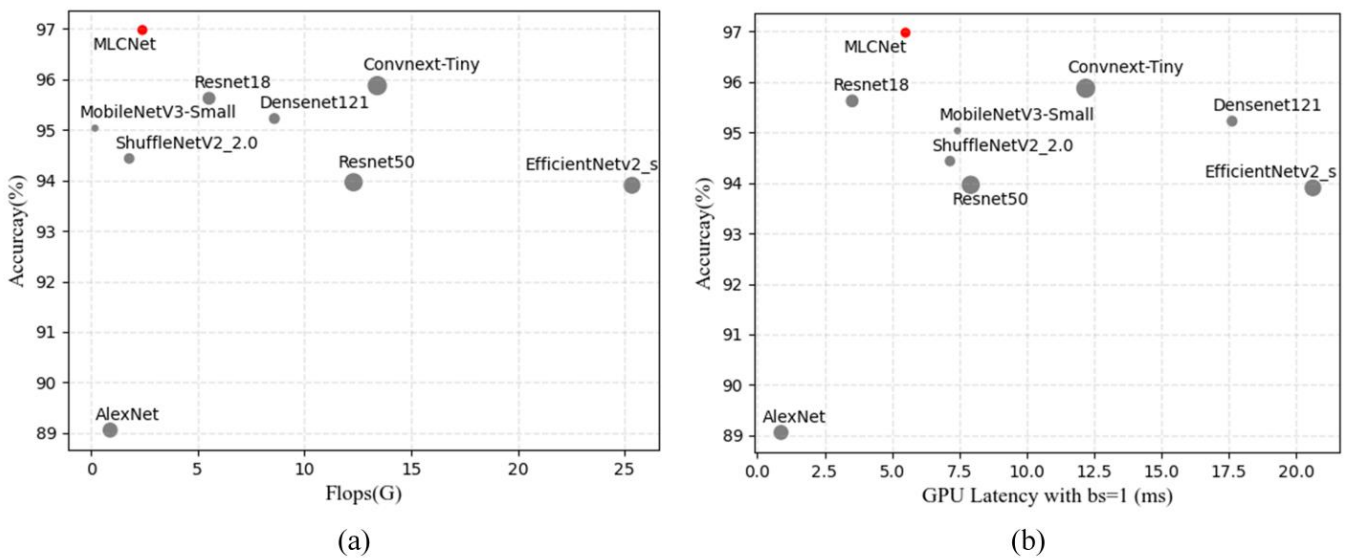

Fig. 8. The classification results of MLCNet and other models. The size of the circle indicates the number of parameters for each model.

*F. Comparison with Existing Multi-view Methods*

There is a lack of previous studies utilizing multi-wavelength data for auroral classification. To demonstrate the effectiveness of our approach in this regard, we conducted a comparison with current multi-view methods. Table VII presents the experimental results obtained with our method and the comparative methods. The results demonstrate that MLCNet outperforms the other comparative methods in three-wavelength auroral image classification. Specifically, the Acc metric of MLCNet surpasses the second ranked MFFNet by 3.19%. In addition, MLCNet achieves a higher Avg_Acc and F1 metrics, outperforming MFFNet by 4.64% and 4.32% respectively. It is worth noting that MLCNet achieves these superior results with only 28% of the inference time of MFFNet. Furthermore, although the number of parameters and FLOPs in MLCNet is slightly higher than in MFFNet, the performance gains justify this trade-off.

TABLE VII
CLASSIFICATION RESULTS OF DIFFERENT MULTI-VIEW MODELS

| Method | Acc | Avg_Acc | F1 | Params(M) | FLOPs(G) | Infer(ms) |
| --- | --- | --- | --- | --- | --- | --- |
| MVDRNet_SE [36] | 93.46% | 90.85% | 91.32% | 134.36 | 46.53 | 17.11 |
| MVDRNet_SK [36] | 92.87% | 90.02% | 90.35% | 190.99 | 266.74 | 87.09 |
| ResNet50_DCCA [34] | 83.36% | 77.00% | 77.56% | 38.36 | 57.41 | 7.92 |
| MCFNet [37] | 91.40% | 86.49% | 88.04% | 5.83 | **0.53** | 16.16 |
| MFFNet [37] | 93.78% | 91.22% | 91.63% | 5.83 | 1.53 | 19.88 |
| MLCNet | **96.97%** | **95.86%** | **95.95%** | **4.57** | 1.65 | **5.50** |

## V. CONCLUSION

In this paper, we present MLCNet, a lightweight auroral classification network that pioneers the automatic classification of auroral images using multi-wavelength observations. MLCNet effectively overcomes the limitations associated with single-wavelength images by amalgamating features derived from auroral images acquired at different wavelengths. Our experimental results clearly demonstrate the improved effectiveness of auroral classification when multi-wavelength images are used, as opposed to relying solely on single-wavelength images. This study not only validates the viability of multi-wavelength auroral image classification, but also introduces new possibilities for automating auroral classification.

In addition, we show that the use of auroral images at 630.0 nm gives better results than the 557.7 nm wavelength commonly used in previous studies. This finding contributes to improving the accuracy of auroral classification by using wavelength-specific observations. It is worth noting that our framework does not impose any special hardware requirements on the observing system. Better image quality naturally leads to better classification results, as is the case with manual analysis. The image size, whether $1024 \times 1024$ or $512 \times 512$, is not an obstacle as it can be scaled using appropriate techniques. Furthermore, our method is adaptable to images taken at other wavelengths, such as 432.0 nm, 540.0 nm or 620.0 nm.

In terms of network performance, the proposed architecture, LCTNet, improves efficiency while maintaining high classification accuracy. The inclusion of the multi-scale reconfiguration module (MSRM) strengthens the network's ability to extract features at different scales. Additionally, the designed LAFE attention module improves the interaction between global and local features, thereby enhancing their discriminative properties. Overall, MLCNet outperforms other auroral classification networks and is superior to previous multi-view approaches. This highlights its usefulness in auroral research.

In the future, we will explore novel fusion methods that can effectively assign importance to each wavelength for different auroral types, thereby improving the accuracy of auroral classification and gaining deeper insights into the characteristics of each auroral type.